# ORCA: An Agentic Reasoning Framework for Hallucination and Adversarial Robustness in Vision-Language Models

Chung-En Johnny Yu

University of West Florida, Pensacola, Florida, USA, cy31@students.uwf.edu

Brian Jalaian

University of West Florida, Pensacola, Florida, USA, bjalaian@uwf.edu

Nathaniel D. Bastian

United States Military Academy, West Point, New York, USA, nathaniel.bastian@westpoint.edu

Large Vision-Language Models (LVLMs) exhibit strong multimodal capabilities but remain vulnerable to hallucinations from intrinsic errors and adversarial attacks from external exploitations, limiting their reliability in real-world applications. We present **ORCA**, an agentic reasoning framework that improves the factual accuracy and adversarial robustness of pretrained LVLMs through inference-time structured inference reasoning with a suite of small vision models ( $\leqq$ 3B parameters). ORCA operates via an *Observe–Reason–Critique–Act* loop, querying multiple visual tools with evidential questions, validating cross-model inconsistencies, and refining predictions iteratively without access to model internals or retraining. ORCA also stores intermediate reasoning traces, which supports auditable decision-making. Though designed primarily to mitigate object-level hallucinations, ORCA also exhibits emergent adversarial robustness without requiring adversarial training or defense mechanisms. We evaluate ORCA across three settings: (1) clean images on hallucination benchmarks, (2) adversarially perturbed images without defense, and (3) adversarially perturbed images with defense applied. On the POPE hallucination benchmark, ORCA improves standalone LVLMs performance by +3.64% to +40.67% across different subsets. Under adversarial perturbations on POPE, ORCA achieves an average accuracy gain of +20.11% across LVLMs. When combined with defense techniques on adversarially perturbed AMBER images, ORCA further improves standalone LVLM performance, with gains ranging from +1.20% to +48.00% across metrics. These results demonstrate that ORCA offers a promising path toward building more reliable and robust multimodal systems.

## 1 INTRODUCTION

Vision-language models (VLMs) have made rapid progress in image captioning, visual question answering (VQA), and multimodal reasoning. Among them, Large Vision-Language Models (LVLMs) extend Vision-Language Pretrained (VLP) models (e.g., CLIP [21], BLIP [13]) with Large Language Models (LLMs), allowing them to generate semantically rich descriptions and multi-step inferences across modalities [4] However, despite their advancements, LVLMs often hallucinate non-existent visual content due to intrinsic model errors or succumb to external adversarial image perturbations. These limitations are especially problematic in high-stakes domains such as remote sensing or medical diagnosis, where factuality and robustness are paramount [2].

Robustness in VLMs is multifaceted. First, intrinsic model errors remain widespread in state-of-the-art LVLMs, especially at the object level, where models confidently assert the existence of objects that are not present. Such hallucinations often stem from biased priors or learned co-occurrence patterns [4]. Second, the recent adversarial attacks targeting multimodal embeddings can cause models to produce confidently wrong predictions using imperceptible image or text perturbations. These attacks are often transferable across architectures, exposing a broad shared vulnerability in VLMs [2, 32]. Importantly, hallucinations and adversarial failures are interrelated, particularly undesirable when compromised visual input is amplified by erroneous model inference [12]. Existing methods attempt to address these challenges through hallucination-aware decoding [18], adversarial training [17], or post-hoc correction [19]. Yet the existing methods are limited in scope, incur high training cost, require model access, or rely on static inference pipelines that cannot adapt to unexpected failures.

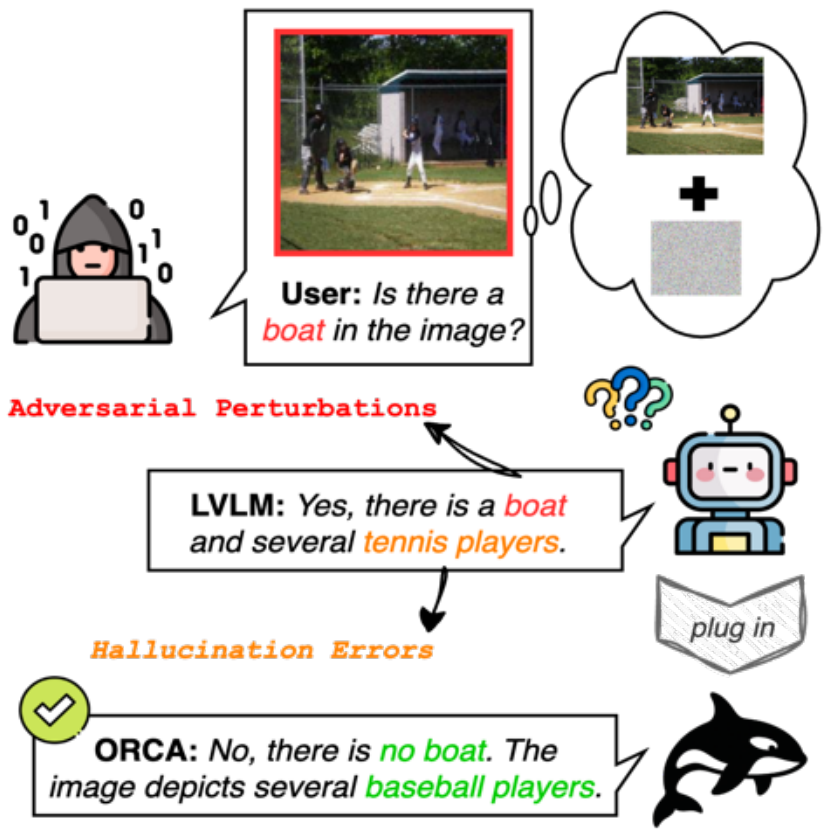

Figure 1. Adversarial perturbations can cause LVLMs to assert nonexistent objects injected by an attacker, and LVLMs hallucinate by their intrinsic errors, whereas ORCA corrects both.

We propose **ORCA** (**O**bserve–**R**eason–**C**ritique–**A**ct), an agentic reasoning framework designed to improve the robustness of LVLMs against hallucination and adversarial perturbations, as illustrated in Figure 1. ORCA operates at test time through an iterative Observe–Reason–Critique–Act loop, querying the baseline LVLM and a suite of lightweight vision models to cross-validate predictions and resolve inconsistencies. Without requiring access to model weights or retraining, a ReAct-style [28] LLM-based agent dynamically generates evidential questions, triangulates answers across tools, and refines its decisions in a structured manner. While originally designed for object-level hallucination mitigation, ORCA exhibits emergent adversarial robustness through architectural diversity and inconsistency-driven reasoning. Its traceable reasoning logs provide auditable decision-making. Our key contributions are:

- We propose ORCA, an agentic reasoning framework that enhances the robustness of pretrained LVLMs by reducing hallucinations and improving adversarial robustness through a inference-time structured inference and correction process. ORCA integrates feedback from a suite of small vision models (≦3B parameters) via an iterative *Observe–Reason–Critique–Act* loop.
- Unlike prior approaches that rely on fine-tuning or adversarial training, ORCA operates entirely at test time without requiring access to model internals or additional data.

- Empirical results demonstrate that ORCA consistently reduces hallucinations, achieving notable average gains across multiple tested models and benchmarks, and exhibits strong adversarial robustness under a range of attacks, all without requiring any adversarial training or defense mechanisms.

## 2 RELATED WORK

Despite the impressive capabilities of LVLMs, hallucinations remain a persistent failure mode. Among various dehallucination approaches, post-hoc correction aims to reduce such hallucinations without retraining or probing the model by extending the inference-time compute [19]. Post-hoc correction is particularly valuable for settings where compute resources are limited or model weights and architectures are inaccessible. Woodpecker [30] extracts object mentions from generated captions and verifies them using additional VQA and object detection models. LogicCheckGPT [25] improves output consistency by prompting the model with logically entailed variations to detect contradictions. DEHALL [26] integrates scene graphs and commonsense reasoning to check alignment between textual descriptions and image semantics in a symbolic manner. While effective in controlled scenarios, these approaches rely on static inference pipelines, require model-specific designs, and remain untested under adversarial settings.

LVLMs are vulnerable to multimodal adversarial attacks due to their tightly coupled image-text embedding space, which expands the attack surface from image to multiple modalities [11]. The recent attacks like Adversarial Illusions [2], Doubly-UAP [12], and AttackVLM [32] reveal how imperceptible perturbations to images or prompts can trigger semantic failures in downstream inference. Moreover, these attacks are often transferable across VLMs due to similar visual encoder architectures, which broaden the range of potentially compromised VLMs [17].

While some existing defenses, such as adversarial training [33] or adversarial prompt tuning [14], have been proposed, they typically require model access and expensive retraining. Conventional image preprocessing techniques such as JPEG compression [8] and feature squeezing [27] are widely adopted in modern attacks [2, 10] to evaluate their attack success rate. The results often indicate that the preprocessing techniques only provide limited protection. While the recent attack, such as Double-UAP [12], has started applying adversarial perturbations to hallucination benchmarks, existing defenses largely overlook the intersection of hallucination mitigation and adversarial robustness, particularly in cases where adversarial inputs amplify intrinsic model errors.

Agentic frameworks offer an emerging alternative for enhancing model reliability through iterative, tool-augmented reasoning at test time. Recent efforts explore agent-based strategies for improving LVLM overall capabilities [24]. For instance, Critic-V [31] adopts an actor-critic paradigm where a Reasoner generates visual reasoning paths and a Critic iteratively refines them using natural language feedback. Similarly, ARA [20] introduces an agentic retrieval mechanism that activates on uncertain generations, selectively injecting retrieval results to mitigate hallucinations in LVLMs. Although these approaches demonstrate the promise of agents for visual reasoning or hallucination mitigation, they do not consider adversarial robustness while enhancing LVLMs. ORCA advances this line of work by offering a training-free framework that simultaneously mitigates hallucinations and resists adversarial perturbations across diverse LVLMs.

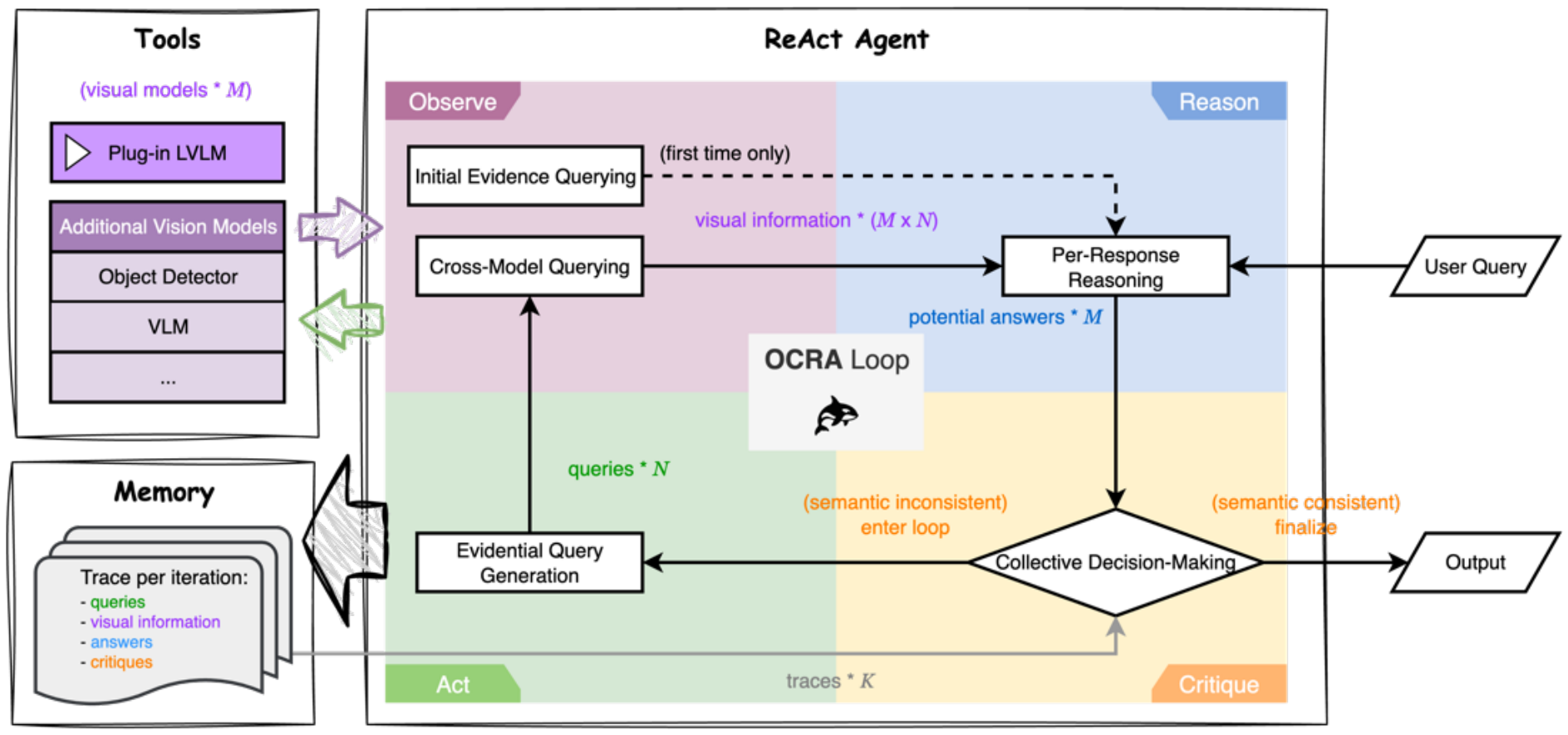

Figure 2. Overview of the ORCA framework. ORCA operates via an Observe–Reason–Critique–Act loop over a plug-in LVLM and a suite of vision tools. At each step, it generates evidential queries, gathers multi-tool responses, cross-validates inconsistencies, and refines predictions, with intermediate reasoning traces logged for auditability.

## 3 METHOD

As illustrated in Figure 2, we implement **ORCA** as a ReAct-style [28] agent operating over pretrained visual models. The agent executes a structured reasoning loop: Observe, Reason, Critique, Act, which iteratively queries vision models to refine its predictions. ORCA comprises three components: an LLM-based agent for reasoning, a set of pretrained vision models (including a plugged-in baseline LVLM), and a memory buffer that logs queries, retrieved evidence, and decisions. The agent interacts with vision tools solely on texts and operates without accessing internal weights.

The framework introduces three key variables (as shown in Figure 2): $M$, the number of visual models; $N$, the number of evidential queries dispatched per iteration; and $K$, the maximum number of iterations (traces) performed before reaching a final decision.

### 3.1 ORCA Reasoning Modules

**Initial Evidence Querying.** Upon receiving a user query, the agent performs *Initial Evidence Querying*, querying the pre-selected visual tools. For example, in Figure 3, it asks the plug-in LVLM to caption and an object detector to detect objects. These models return rich image contents, which serve as the foundational evidence for downstream reasoning. This step is executed once per sample to bootstrap the agent's understanding of the image.

**Per-Response Reasoning.** The agent then evaluates each vision model's response individually in the *Per-Response Reasoning* module by assessing whether the retrieved content supports or contradicts the user query. This forms the $M$ potential answers to the user query corresponding to $M$ vision models. For binary object presence tasks, it first extracts the target object (e.g., "*person*" in Figure 3) from the user query and uses it as the anchor for evaluating alignment and guiding subsequent evidential queries. This reasoning process is conducted entirely in the text domain using the LLM's inherent inferential capabilities, making it more robust to adversarial perturbations that affect image modality.

**Collective Decision-Making and Consistency Check.** After gathering the potential answers, the agent performs *Collective Decision-Making*, aggregating the answers to determine whether there is semantic agreement across vision models. If the agent finds the answers consistent, it finalizes the answer and outputs the result. If inconsistencies are detected, such as conflicting or ambiguous answers, the agent enters the ORCA loop to retrieve additional evidence from visual tools. Consistency checking in ORCA is a symbolic mechanism, where the agent applies rule-based templates (e.g., if-else conditions) that encode human knowledge about visual tools strengths and weaknesses. For instance, as shown in Figure 3, an object detector reports that "*no person is detected*" while an LVLM caption states "*unclear if the frisbee is thrown by a person.*" Given that object detectors can fail under distribution shift, e.g., when a person's appearance deviates from the training distribution, and the LVLM implicitly suggests the presence of a person, this inconsistency triggers entry into the reasoning loop. Such symbolic hooks support auditable logic flow. While not explicitly designed for adversarial defense, ORCA's iterative inconsistency-checking and vision model diversity contribute to its emergent robustness by reducing the vulnerability of a single model.

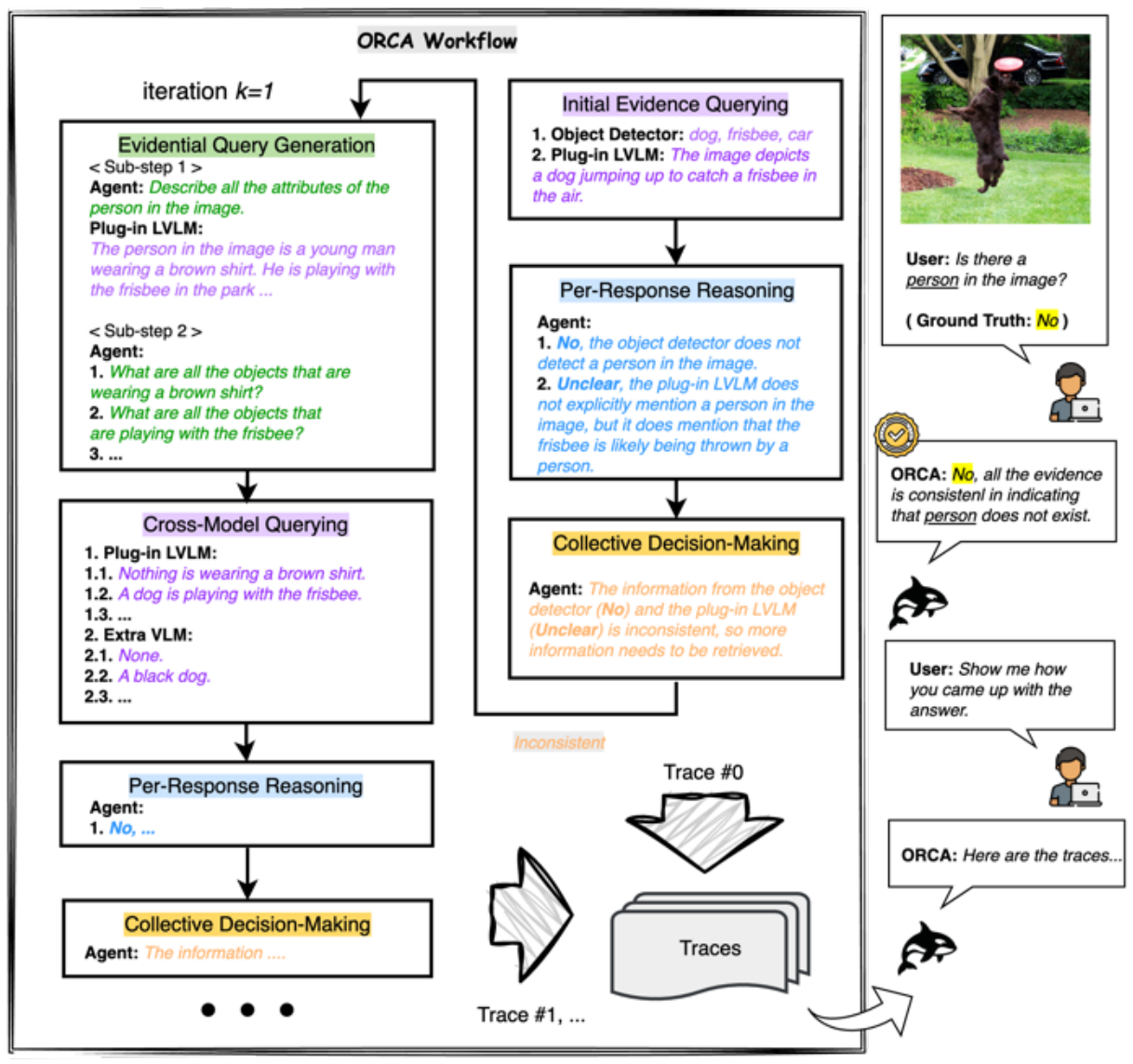

Figure 3. ORCA corrects false predictions from standalone LVLMs by querying diverse vision models and resolving inconsistencies. Reasoning traces show details of ORCA's decision-making process.

**Evidential Query Generation.** To resolve inconsistencies, ORCA employs a strategy of *attribute-guided evidential querying*, which has been proved effective in LogicCheckGPT [25]. When a disagreement arises about the presence of a target object, the agent first prompts the plug-in LVLM to describe the object's attributes (e.g., color, count, location). These descriptions are then used to programmatically construct a set of attribute-based questions designed to cross-verify the existence of the object across models. For instance, if the LVLM states that "*the person is wearing a brown shirt,*" the agent may issue follow-up queries such as "*What objects are wearing a brown shirt?*" The agent generates $N$ such evidential queries per iteration, which are dispatched in parallel to the $M$ visual models. This creates an evidence pool of up to $M \times N$ responses per iteration, significantly increasing the evidence diversity available for cross-checking.

**Cross-Model Querying.** Each evidential query is routed to $M$ vision models, yielding diverse but potentially overlapping responses. By leveraging architectural diversity in visual tools, ORCA spreads the risk of model-specific weakness or adversarial vulnerabilities.

**Loop and Memory Tracing.** The loop continues until a consistent decision is reached or a maximum iteration count $K$ is exceeded. All responses, critiques, and decisions are stored in memory. At the final iteration $K$, if the agent lacks consistent evidence to support any answer, it revisits historical traces to make the final determination. Beyond guiding reasoning, the traces also enable human users to backtrace ORCA's decision-making process.

### 3.2 ORCA Workflow Example

Figure 3 illustrates a concrete example of the iterative ORCA loop when tackling a binary object existence question. When queried about the presence of a person in the image, the agent initially receives conflicting evidence between the object detector and plug-in LVLM. It reflects the inconsistency and enters the ORCA loop, generating evidential queries to gather object existence evidence from multiple vision tools. These responses are then re-evaluated, and once sufficient consistency is reached, the agent finalizes its answer. Importantly, all intermediate steps are stored as traces, enabling auditability of the decision process.

While the ORCA loop is primarily described in the context of binary VQA, the framework also supports image captioning. In our implementation, the agent first queries three VLMs to generate initial captions. From these, it first extracts the overlapping captions among all three models, then extracts a candidate list of potentially existing objects by identifying frequently mentioned entities and contextual overlaps. Each object is then treated as a query target, and the standard ORCA loop is applied to verify its existence using evidential queries and cross-model reasoning. This object-level validation ensures that the final caption is grounded in verifiable visual evidence, rather than relying solely on any single model's generation.

## 4 EXPERIMENT

### 4.1 Experiment Settings

We design a comprehensive experimental setting to evaluate the robustness of the ORCA framework, with a focus on reducing hallucinations and withstanding adversarial attacks across a variety of LVLMs. While ORCA is model-agnostic and modular by design, we instantiate it using commonly available open-source LVLMs and small (≦3B parameters) visual models.

**Hallucination Benchmarks.** To assess dehallucination abilities, we evaluate ORCA on three widely used benchmarks that assess object-level hallucination. The POPE benchmark [15] tests binary object presence questions (e.g., "*Is there a {OBJECT} in the image?*"). It includes three subsets: *Random Subset* randomly chooses absent objects; *Popular Subset* selects frequently occurring but absent objects; and *Concurrence Subset* [1] introduces plausible distractors based on concurrence statistics.

We further incorporate the *Existence Subset* of the MME benchmark [16], which presents binary object presence queries akin to POPE. MME reports Accuracy, Accuracy Plus (Acc+), which is scored by correctly answering a pair of questions on the same image, and Total Score, the sum of Acc and Acc+.

For captioning hallucination evaluation, we adopt the *Generative Subset* from the AMBER benchmark [23], where models generate image captions and are evaluated based on the inclusion of predefined object lists: both ground-truth objects and hallucinated distractors. We report three standard metrics from the benchmark. *CHAIR* measures the frequency of hallucinated object mentions by computing the proportion of objects in a response that do not appear in the ground-truth object list; a lower CHAIR score indicates fewer hallucinations. *Hal* represents the proportion of responses that contain any hallucination, i.e., it assigns 1 to a response if CHAIR is non-zero, and 0 otherwise, capturing the overall prevalence of hallucination. *Cog* assesses the proportion of generated objects that match a predefined set of cognitively plausible hallucinations, quantifying how closely the model's mistakes resemble typical human perceptual errors. We omit the *Cover* metric, which measures ground-truth object coverage, as ORCA is designed to suppress hallucinations rather than maximize object recall. Due to the high computational cost of downstream adversarial evaluation, we subsample 50 images from AMBER.

**Adversarial Settings.** To stress-test robustness under external perturbations, we follow prior work [12] and apply two modern adversarial attacks to the images in the hallucination benchmarks. Both attacks exhibit transferability across

1 *Concurrence Subset* originally is named as *Adversarial Subset*. To avoid confusion, we renamed it to separate it from the adversarial attacks. These subsets represent different perspectives of hallucinations. We evaluate 300 questions per subset.

VLMs, meaning not only the baseline LVLM is targeted, but also the additional vision models in ORCA. This places ORCA at a disadvantage compared to single-model baselines. The first attack, Adversarial Illusions [2], is a white-box method that uses ImageBind [9] as the surrogate model to inject non-existent objects into multimodal embeddings, with a perturbation budget of $\varepsilon = 16/225$. The second, [32], is a black-box, query-based attack (MF-ii + MF-tt) that uses UniDiffuser [5] as the surrogate to generate mismatched image-text pairs, with a budget of $\varepsilon = 8/225$. We adopt the default configurations for as defined in their original works, which have been shown to be effective.

In addition to adversarial attacks, we apply two traditional pixel-level defenses to the benign and perturbed AMBER images: (1) JPEG compression [8] (quality = 50), which reduces high-frequency perturbations while preserving semantic content; and (2) feature squeezing [27] (bit depth = 4), which quantizes pixel values to suppress adversarial noise. These defenses are lightweight, training-free, and widely adopted for their simplicity and efficiency. By testing these universal defenses, we aim to assess ORCA's compatibility with practical robustness tools that can be readily applied in black-box settings. We evaluate their impact on both standalone LVLMs and ORCA under adversarial conditions, using standard parameter settings as in [10].

**Implementation Details.** We instantiate ORCA independently with each of three widely adopted LVLMs: mPLUG-Owl (7B) [29], MiniGPT-4 (13B) [34], and Qwen-VL-Chat (13B) [3], which serve as both plug-in models and standalone baselines. These models span diverse vision encoder backbones: CLIP, EVA, and OpenCLIP respectively, which provides architectural diversity that further validates ORCA's generalizability. To complement these, we integrate three additional models that offer diverse architectures and visual perspectives: DETR (42M) [7] as an efficient object detector, Paligemma (3B) [6] as a compact VLM, and BLIP (385M) [13] for VQA and captioning. The agent's backbone is LLaMA-3.2-Vision-Instruct (11B) [1], which processes only text and does not access raw images. That is, all agent actions are made solely based on textual responses retrieved from the vision models. ORCA performs a maximum of three iterations per example.

Table 1: Performance on POPE. Each LVLM is evaluated across three subsets: Random, Popular, and Concurrence. **Bold** values indicate the highest score for each metric within a LVLM.

| **LVLM** | **Method** | **Random** | | **Popular** | | **Concurrence** | |
|---|---|---|---|---|---|---|---|
| | | Acc | F1 | Acc | F1 | Acc | F1 |
| mPlug-Owl | Standalone | 54.00 | 68.49 | 51.33 | 67.26 | 51.00 | 67.11 |
| | + ORCA | **94.67** | **94.37** | **92.00** | **91.84** | **90.33** | **90.24** |
| MiniGPT-4 | Standalone | 75.00 | 76.19 | 72.00 | 75.15 | 70.00 | 74.29 |
| | + ORCA | **93.67** | **93.38** | **93.33** | **93.01** | **92.00** | **91.84** |
| Qwen-VL-Chat | Standalone | 90.33 | 89.90 | 86.00 | 86.00 | 85.67 | 85.71 |
| | + ORCA | **94.00** | **93.62** | **92.67** | **92.31** | **92.33** | **91.93** |

### 4.2 Results Analysis

**Robustness to Hallucinations.** Across all three POPE subsets: Random, Popular, and Concurrence, ORCA consistently improves both accuracy and F1 scores for all LVLMs, as shown in Table 1. For example, ORCA raises mPLUG-Owl's accuracy from 54.00% to 94.67% in the Random subset, a gain of over 40%. Similar gains are observed in the Popular and Concurrence subsets, with ORCA narrowing the performance gap even in more challenging hallucination scenarios. For instance, in the Popular subset, MiniGPT-4 improves from 72.00% to 93.33%. The standalone Qwen-VL-Chat drops 4% (from 90.33% in Random subset to 85.67% in Concurrence subset), whereas ORCA stabilizes within a 2% margin, maintaining over 92% accuracy across all subsets. Overall, ORCA improves standalone LVLMs performance by +3.64% to +40.67% across different subsets.

On the MME benchmark (Existence Subset), ORCA de-hallucinates all the LVLMs, as shown in Table 2. Regardless of the LVLM, ORCA consistently reaches a Total Score of 180.00, with both Accuracy and Accuracy+ improving to 93.33% and 86.67%, respectively. Even weaker baselines like mPLUG-Owl benefit significantly, doubling their performance compared to standalone operation. This demonstrates ORCA's generalizability across structurally similar hallucination tasks.

Table 2: Performance on MME. **Bold** values indicate the highest score for each metric within a LVLM.

| LVLM | Method | Acc | Acc+ | Total |
|---|---|---|---|---|
| mPlug-Owl | Standalone | 63.33 | 26.67 | 90.00 |
| | + ORCA | **93.33** | **86.67** | **180.00** |
| MiniGPT-4 | Standalone | 66.67 | 50.00 | 116.67 |
| | + ORCA | **93.33** | **86.67** | **180.00** |
| Qwen-VL-Chat | Standalone | 91.67 | 83.33 | 175.00 |
| | + ORCA | **93.33** | **86.67** | **180.00** |

For hallucination evaluation on captioning task, the AMBER Generative Subset in Table 3 shows consistent de-hallucination effects across all metrics for all LVLMs. For example, MiniGPT-4's CHAIR score drops from 16.30% to 8.70%, and Hal from 84.00% to 26.00%, which is a 58% drop. Qwen-VL-Chat's Cog score decreases from 4.10% to just 0.40%. While CHAIR and Hal reductions reflect suppression of false object mentions, the steep drop in Cog further highlights ORCA's ability to filter out contextually plausible but incorrect generations. Together, these results validate ORCA as a model-agnostic de-hallucination framework that mitigates hallucinations on both discriminative and generative tasks.

Table 3: Results on the AMBER (benign setting). Lower values indicate better performance. **Bold** values indicate the best (lowest) score for each metric within a LVLM.

| LVLM | Method | CHAIR ↓ | Hal ↓ | Cog ↓ |
|---|---|---|---|---|
| mPlug-Owl | Standalone | 23.20 | 82.00 | 16.70 |
| | + ORCA | **19.70** | **56.00** | **7.70** |
| MiniGPT-4 | Standalone | 16.30 | 84.00 | 12.60 |
| | + ORCA | **8.70** | **26.00** | **1.60** |
| Qwen-VL-Chat | Standalone | 7.50 | 40.00 | 4.10 |
| | + ORCA | **3.20** | **8.00** | **0.40** |

**Adversarial Robustness.** We first assess adversarial robustness across attack types by averaging accuracy across the three POPE subsets under benign, Adversarial Illusions, and AttackVLM settings, as shown in Figure 4; numerical results in Appendix Table 7. For all three LVLMs, ORCA significantly outperforms their standalone versions across all settings. For instance, mPLUG-Owl sees a jump from 52.11% to 92.33% on clean inputs with ORCA, and maintains 85.00% and 87.00% accuracy under Adversarial Illusions and AttackVLM. Similar patterns are observed with MiniGPT-4 and Qwen-VL-Chat; ORCA achieves an average accuracy gain of +20.11% across LVLMs. This confirms ORCA's ability to preserve performance under both white-box and black-box attacks.

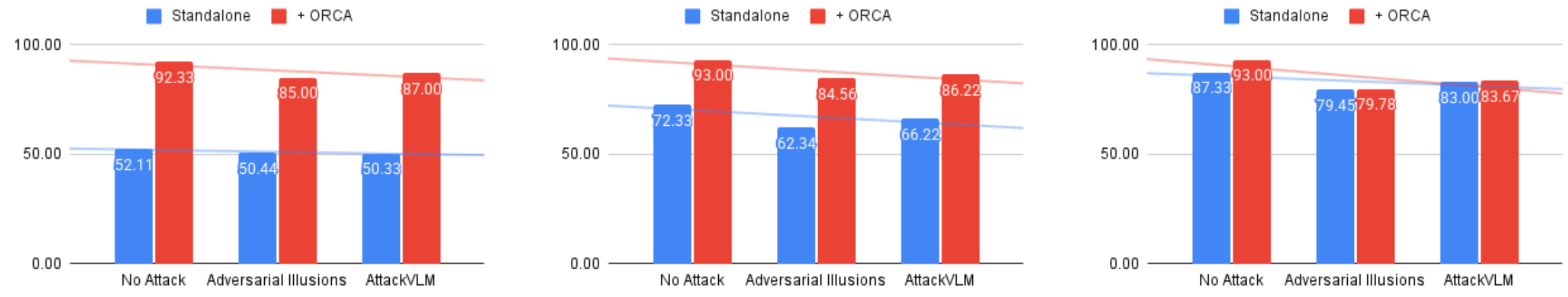

(a) mPlug-owl (b) MiniGPT-4 (c) Qwen-VL-Chat

Figure 4. Average accuracy across the three subsets of POPE, comparing standalone LVLMs and ORCA-augmented versions across adversarial conditions: No Attack, Adversarial Illusions, and AttackVLM.

To further examine robustness across LVLMs under a fixed attack, radar plots in Figure 5 visualize the averaging accuracy across the three POPE subsets. Under both Adversarial Illusions and AttackVLM, ORCA consistently boosts each LVLM's accuracy above 79%, highlighting its generalizable defense capability regardless of the model architecture. The numerical results are in Appendix Table 8.

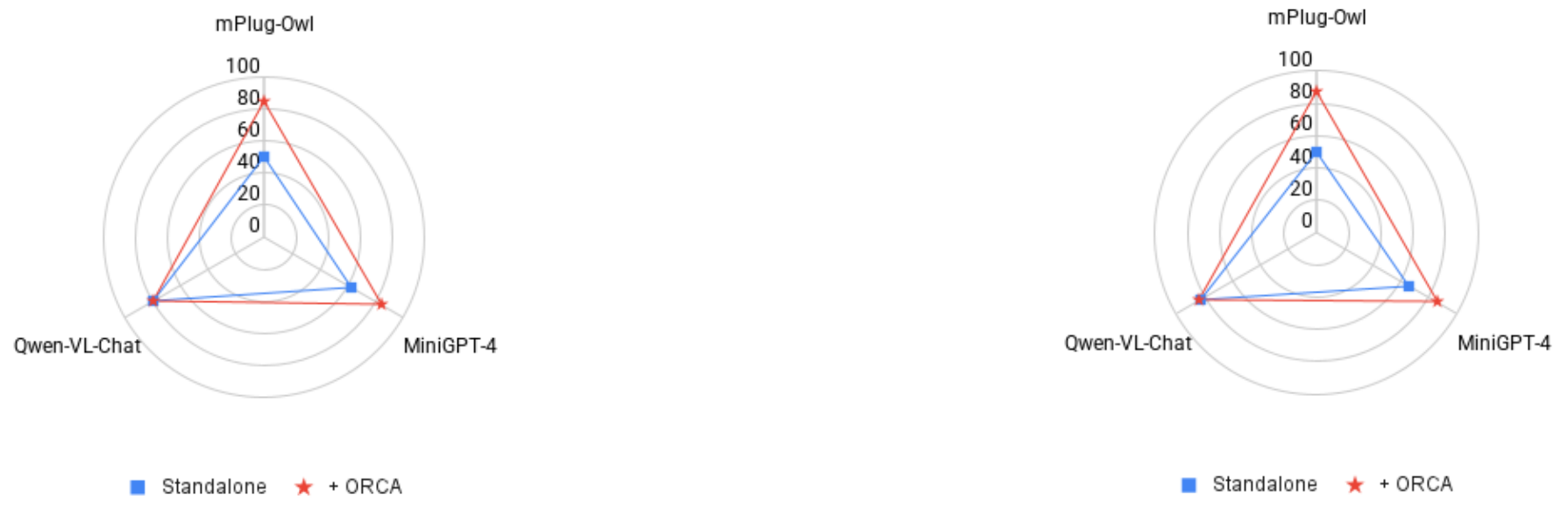

(a) Adversarial Illusions (b) AttackVLM

Figure 5. Performance comparison under two attack settings. Each vertex represents one LVLM, and the score reflects average accuracy across the three subsets of POPE. Refer to Appendix Table 7, 8 for full numerical results of each individual subset.

We also validate these findings on the AMBER Generative Subset using adversarially perturbed images, as shown in Table 4 and 5. For example, with Adversarial Illusions, MiniGPT-4's mPLUG-Owl's CHAIR decreases from 24.50% to 18.80%, and Hal score drops from 74.00% to 38.00%. Qwen-VL-Chat's Cog score falls from 3.30% to 0.40%. Comparable improvements are seen under AttackVLM, reinforcing ORCA's robustness on both closed-form and open-ended tasks under adversarial conditions.

Table 4: Results on adversarially perturbed AMBER (Adversarial Illusions). Lower values indicate better performance. **Bold** values indicate the best (lowest) score for each metric within each LVLM.

| Adversarial Illusions | | | | |
|---|---|---|---|---|
| **LVLM** | **Method** | **CHAIR ↓** | **Hal ↓** | **Cog ↓** |
| mPlug-Owl | Standalone | 24.50 | 70.00 | 10.60 |
| | + ORCA | **18.80** | **40.00** | **4.50** |
| MiniGPT-4 | Standalone | 17.70 | 74.00 | 11.40 |
| | + ORCA | **14.50** | **38.00** | **1.60** |
| Qwen-VL-Chat | Standalone | 10.40 | 34.00 | 3.30 |
| | + ORCA | **9.30** | **24.00** | **0.40** |

Table 5: Results on adversarially perturbed AMBER (AttackVLM). Lower values indicate better performance. Bold values highlight the best (lowest) scores within each LVLM.

| AttackVLM | | | | |
|---|---|---|---|---|
| **LVLM** | **Method** | **CHAIR ↓** | **Hal ↓** | **Cog ↓** |
| mPlug-Owl | Standalone | 24.40 | 84.00 | 11.80 |
| | + ORCA | **20.70** | **54.00** | **4.90** |
| MiniGPT-4 | Standalone | 16.40 | 78.00 | 13.40 |
| | + ORCA | **13.50** | **40.00** | **3.70** |
| Qwen-VL-Chat | Standalone | 7.00 | 30.00 | 3.30 |
| | + ORCA | **4.50** | **10.00** | **0.00** |

**Impacts of Adversarial Defense.** We evaluate whether universal pixel-level defenses improve robustness on adversarial AMBER images. While defenses improve the performance of standalone LVLMs, ORCA consistently outperforms its corresponding baselines across all settings. The results are illustrated in Table 6. For example, with JPEG compression, MiniGPT-4's CHAIR score drops from 21.00% to 9.90%, Hal from 72.00% to 24.00%, and Cog from 14.20% to 1.40% when applying ORCA. Similarly with feature squeezing, mPlug-Owl's Hal drops significantly from 70% to 32% when applying ORCA. Overall, ORCA further improves standalone LVLM performance, achieving additional performance gains ranging from +1.20% to +48.00% across metrics.

Table 6: Results on Adversarial Illusions across defenses and LVLMs on AMBER. Lower values indicate better performance. **Bold** values denote the best score under each defense and metric.

| Adversarial Illusions under Defenses | | | | | |
|---|---|---|---|---|---|
| **LVLM** | **Defense** | **Method** | **CHAIR ↓** | **Hal ↓** | **Cog ↓** |
| mPlug-Owl | No defense | Standalone | 24.50 | 70.00 | 10.60 |
| | | + ORCA | **18.80** | **40.00** | **4.50** |
| | JPEG Compress | Standalone | 18.50 | 54.00 | 9.30 |
| | | + ORCA | **20.70** | **40.00** | **4.90** |
| | Feature Squeeze | Standalone | 19.50 | 70.00 | 10.20 |
| | | + ORCA | **13.40** | **32.00** | **3.30** |
| MiniGPT-4 | No defense | Standalone | 17.70 | 74.00 | 11.40 |
| | | + ORCA | **14.50** | **38.00** | **1.60** |
| | JPEG Compress | Standalone | 21.00 | 72.00 | 14.20 |
| | | + ORCA | **9.90** | **24.00** | **4.10** |
| | Feature Squeeze | Standalone | 15.60 | 60.00 | 9.80 |
| | | + ORCA | **8.40** | **22.00** | **1.20** |
| Qwen-VL-Chat | No defense | Standalone | 10.40 | 34.00 | 3.30 |
| | | + ORCA | **9.30** | **24.00** | **0.40** |
| | JPEG Compress | Standalone | 10.90 | 34.00 | 1.60 |
| | | + ORCA | **5.80** | **16.00** | **0.40** |
| | Feature Squeeze | Standalone | 12.80 | 42.00 | 3.30 |
| | | + ORCA | **10.10** | **26.00** | **1.20** |

## 5 DISCUSSION

**Why Agentic Reasoning Helps.** Our results demonstrate that agentic reasoning, when structured through the ORCA framework, substantially improves the robustness of vision-language systems by increasing the inference-time compute without retraining or internal model access. ORCA operates with small pretrained vision models (≦3B parameters), yet outperforms standalone LVLMs by leveraging architectural diversity and structured inconsistency-driven querying. Its effectiveness against both intrinsic hallucinations and external perturbations suggests that robustness arises not solely from model scale, but from the design of interaction and reasoning mechanisms. We posit that combining model diversity with iterative cross-model validation mitigates hallucinations and enhances adversarial robustness.

**Beyond Object-Level Hallucinations.** While this work focuses on object-level hallucinations, ORCA's reasoning paradigm naturally extends to more complex forms of hallucination—such as attributes and relations. These hallucinations can potentially be addressed by ORCA's evidential querying on specialized vision models. By dynamically validating and revising predictions based on diverse visual tools, ORCA provides a flexible foundation for tackling fine-grained multimodal reasoning challenges beyond object presence alone.

**Validating the Adversarial Setting.** Our evaluation includes two widely adopted adversarial attacks: AttackVLM [32], which has demonstrated strong transferability across models such as MiniGPT-4 [34] and BLIP [13], and Adversarial Illusions [2], which is designed to bypass conventional defenses like JPEG compression [8] and feature squeezing [27]. While these attacks may not represent the most optimized threat against ORCA specifically, their use reflects a stronger threat model, where transfer attacks simultaneously target multiple vision models within ORCA. The fact that ORCA maintains strong performance under these conditions highlights the robustness of its reasoning architecture and its ability to benefit from architectural diversity among vision tools.

While the empirical results demonstrate ORCA's robustness against adversarial perturbations without any defenses, we observe that ORCA with conventional defenses can further improve adversarial robustness. Unlike adversarial training or prompt tuning, which require model access and significant compute, ORCA offers a practical, model-agnostic alternative that enhances robustness at test time through structured inconsistency-aware reasoning. This makes ORCA particularly appealing for real-world scenarios where model internals are inaccessible or retraining is infeasible.

**Toward Reliable and Extensible Multimodal AI.** ORCA advances robustness in vision-language systems by enabling factual reliability and adversarial robustness through structured reasoning and model diversity without requiring access to model internals or retraining. While its current implementation incurs higher inference-time latency (e.g., $\approx 33\times$ slower than standalone LVLMs on POPE), this trade-off is acceptable in high-stakes domains such as medical diagnosis or remote sensing, where reliability is paramount. Also, our implementation is not optimized. Moreover, ORCA's modular design allows for plug-and-play integration of faster or more specialized models, and efficiency can be improved via convergence-based stopping or adaptive routing. Looking forward, ORCA's flexibility makes it amenable to integration with robustness-enhancing components, such as adversarially fine-tuned models like Robust CLIP [22] or domain-specific tools. Its transparent decision process offers an interpretable reasoning trace, supporting accountability in critical applications. Future work should also address limitations, such as confidently incorrect cross-model agreement, to further improve trustworthiness and generalizability in agentic multimodal systems.

## 6 CONCLUSION

We presented ORCA, a inference-time agentic reasoning framework that enhances the robustness of pretrained LVLMs against both hallucination and adversarial perturbations. ORCA operates through a structured *Observe–Reason–Critique–Act* loop, leveraging a suite of lightweight vision models ($\leqq$3B parameters) to cross-validate predictions and iteratively refine decisions. It does not require model access, fine-tuning, or additional training data. Our experiments demonstrate that ORCA consistently reduces object-level hallucinations across multiple LVLMs and benchmarks and exhibits strong robustness under a range of adversarial attacks. Notably, these robustness gains emerge without the use of adversarial training or defense-specific techniques, highlighting the power of architectural diversity and inconsistency-driven reasoning. Furthermore, ORCA produces auditable reasoning traces, making it a compelling direction for reliable multimodal AI systems.

## ACKNOWLEDGMENTS

This work was supported in part by the U.S. Military Academy (USMA) under Cooperative Agreement No. W911NF-23-2-0108. The views and conclusions expressed in this paper are those of the authors and do not reflect the official policy or position of the U.S. Military Academy, U.S. Army, U.S. Department of Defense, or U.S. Government.

# A APPENDICES

## A.1 Additional Results

Table 7 and Table 8 provide expanded breakdowns of the main results presented in Section 3.

Table 7: Performance on POPE under attack of Adversarial Illusions. Each LVLM is evaluated across three subsets: Random, Popular, and Concurrence. **Bold** values indicate the highest score for each metric within a model and subset.

| Adversarial Illusions | | | | | | | |
|---|---|---|---|---|---|---|---|
| **LVLM** | **Method** | **Random** | | **Popular** | | **Concurrence** | |
| | | Acc | F1 | Acc | F1 | Acc | F1 |
| mPlug-Owl | Standalone | 51.33 | 67.26 | 50.00 | 66.67 | 50.00 | 66.67 |
| | + ORCA | **86.00** | **84.44** | **87.67** | **86.64** | **81.33** | **80.00** |
| MiniGPT-4 | Standalone | 66.67 | 74.09 | 58.67 | 69.00 | 61.67 | 71.18 |
| | + ORCA | **85.00** | **83.14** | **88.00** | **86.67** | **80.67** | **79.58** |
| Qwen-VL-Chat | Standalone | 79.67 | 76.45 | 79.00 | 75.49 | **79.67** | **76.62** |
| | + ORCA | **80.33** | **75.92** | **82.00** | **78.40** | 77.00 | 72.06 |

Table 8: Performance on POPE under attack of AttackVLM. Each LVLM is evaluated across three subsets: Random, Popular, and Concurrence. **Bold** values indicate the highest score for each metric within a model and subset.

| AttackVLM | | | | | | | |
|---|---|---|---|---|---|---|---|
| **LVLM** | **Method** | **Random** | | **Popular** | | **Concurrence** | |
| | | Acc | F1 | Acc | F1 | Acc | F1 |
| mPlug-Owl | Standalone | 51.00 | 67.11 | 50.00 | 66.67 | 50.00 | 66.67 |
| | + ORCA | **88.67** | **87.50** | **86.33** | **84.98** | **86.00** | **85.21** |
| MiniGPT-4 | Standalone | 71.67 | 77.21 | 62.67 | 72.28 | 64.33 | 73.58 |
| | + ORCA | **86.33** | **84.64** | **87.67** | **86.35** | **84.67** | **83.57** |
| Qwen-VL-Chat | Standalone | 83.67 | 80.48 | **83.00** | 79.84 | 82.33 | 79.21 |
| | + ORCA | **84.33** | **81.57** | **83.00** | **80.16** | **83.67** | **80.93** |

## A.2 Prompt Templates

We provide the prompt templates used in our experiments for reproducibility, as shown in Figure 6, 7, 8, 9. We include only the most essential templates that were central to our experiments, omitting minor variations for brevity.

**System Prompt**
You are a helpful language assistant that helps to generate a question according to instructions.

---

**Prompt**
You will receive a piece of text that describes an object, and the given object.
**[Task]**
Your task is to accurately identify and extract every attribute associated with the given object in the provided text. Each claim should be concise (less than 15 words) and self-contained, corresponding to only one attribute. You MUST only respond in the format as required. Each line should contain the original claim and the modified claim with all the mentions of the given object being replaced with ”the object”. DO NOT RESPOND WITH ANYTHING ELSE. ADDING ANY OTHER EXTRA NOTES THAT VIOLATE THE RESPONSE FORMAT IS BANNED.

**[Response Format]**
original claim&modified claim

Here are examples:
{examples}
Now complete the following:
**[Text]:**
{sent}
**[Entity]:**
{entity}
**[Response]:**

Figure 6. Prompt template: Extracting attributes for generating evidential queries, which is referred from LogicCheckGPT [25].

**System Prompt**

You are a language assistant that helps to rephrase sentences from given sentences.

---

**Prompt**

You will receive a list of statements of objects in an image.
**[Task]**
Your task is to rephrase each line of statement into a question following the below question template.
In specific, extract attributes of the object to fill in the attribute slot of the template to form questions.
DO NOT RESPOND WITH ANYTHING ELSE. DO NOT CHANGE THE QUESTION TEMPLATE.

**[Template]:**
What are all the objects that (ATTRIBUTE SLOT) in the image?

**[Response Format]:**
What are all the objects that are yellow in the image?
What are all the objects that are parked near a marketplace in the image?
What are all the objects that are made of wood in the image?
What are all the objects that are walking down the street in the image?
What are all the objects that have tablecloths on them on the street in the image?

Now complete the following:
**[Statements]:**
{statement}
**[Response]:**

Figure 7. Prompt template: Generating evidential queries, which is referred from LogicCheckGPT [25].

**System Prompt**
You are a language assistant that helps to answer the question according to instructions.

---

**Prompt**
You are given information and a question. The information is the caption of an image.

**[Task]**
Your task is to answer the question according to the given information and provide step by step reasoning. No matter the question asks you to answer 'Yes' or 'No', your answer can be 'Yes', 'No', or 'Unclear'. Your answer should plausible and consistent with your reasoning. Your reasoning should be based on the given information and think it through step by step. If the information mentions that the object in the question exists in the image, your answer should be Yes. If the information does not explicitly mention the object in the question, but the mentioned objects imply the object in the question exists, or the object in the question is typically expected in the described scene, your answer should be Unclear. If the information does not explicitly mention the object in the question, and mentioned objects does not imply the object in the question exists, and the object in the question is not typically expected in the described scene, your answer should be No. All the objects can be inferred with exact matches, synonyms, plural forms, or well-defined subclass relationships. Do not infer loosely related objects. Please follow the output format. DO NOT RESPOND WITH ANYTHING ELSE. ADDING ANY OTHER EXTRA NOTES THAT VIOLATE THE OUTPUT FORMAT IS BANNED.

**[Output Format]**
Possible Answer: (your answer)
Reasoning: (your reasoning)

Now complete the following:
**[Information]**
{information}
**[Question]**
{question}
**[Output]**

Figure 8. Prompt template: Per-response reasoning example of recieving LVLM's caption.

**Rule Description**

Determine a final decision from object detector and captioning-based VLM outputs, using symbolic consistency logic that encodes human trust priors over different vision tools.

**Decision Logic**

`If detector = "Yes"` and `caption = any`, **return "Yes"**
`If detector = "No"` and `caption = "No"`, **return "No"**
Otherwise, **return "Unclear"**

**Design Rationale**

Object detectors are prioritized due to stronger visual grounding. They are less prone to hallucinate objects that do not exist. When a detector returns "Yes", we trust its localization ability even if the captioning model disagrees. If the detector returns "No", we require mutual agreement from the captioner to conclude "No"; otherwise, we fall back to "Unclear" to account for uncertainty or false negatives.

Figure 9. Cross-model consistency rule example: A symbolic decision module for fusing predictions from object detection and LVLM captioning.

## Authors' background

| Your Name | Title* | Research Field | Personal website |
|---|---|---|---|
| Chung-En Johnny Yu | PhD Student | AI Agent Systems , Reliable Multimodal AI | |
| Brian Jalaian | Associate Professor | | |
| Nathaniel D. Bastian | Assistant Professor | | |

*This form helps us to understand your paper better, the form itself will not be published.

*Title can be chosen from: master student, Phd candidate, assistant professor, lecture, senior lecture, associate professor, full professor